%% file: main.tex
\documentclass[conference]{IEEEtran}
\IEEEoverridecommandlockouts
\usepackage{cite}
\usepackage{amsmath,amssymb,amsfonts}
\usepackage{algorithmic}
\usepackage{graphicx}
\usepackage{textcomp}
\usepackage{multirow}
\usepackage{booktabs}
\usepackage{makecell}
\usepackage{xcolor}
\usepackage{hyperref}
\usepackage{url}
\def\BibTeX{{\rm B\kern-.05em{\sc i\kern-.025em b}\kern-.08em
    T\kern-.1667em\lower.7ex\hbox{E}\kern-.125emX}}

\makeatletter
\newcommand{\twocolumnfootnotefullwidth}[1]{%
  \begingroup
  \renewcommand{\thefootnote}{}
  \footnotetext{%
    \noindent\hspace*{-1em}\rule{0.3\linewidth}{0.4pt}\\[0.2em] 
    \vspace*{-1em}
    \noindent\footnotesize #1
  }
  \endgroup
}
\makeatother

\begin{document}

\title{CIPHER: Counterfeit Image Pattern High-level Examination via Representation for GAN and Diffusion Discriminator Learning}


\author{%
\makebox[\textwidth][c]{%
\begin{tabular}{ccccc}
Kyeonghun Kim & Youngung Han\textsuperscript{*} & Seoyoung Ju & YeonJu Jean & YooHyun Kim \\
\textit{OUTTA} & \textit{OUTTA} & \textit{OUTTA} & \textit{OUTTA} & \textit{OUTTA} \\
kyeonghun.kim@outta.ai & youngung.han@outta.ai & seoyoung.ju@outta.ai & yeonju.jean@outta.ai & yoohyun.kim@outta.ai
\end{tabular}}%
\\[1.2ex]
\makebox[\textwidth][c]{%
\begin{tabular}{ccccc}
Minseo Choi & SuYeon Lim & Kyungtae Park & Seungwoo Baek & Sieun Hyeon\textsuperscript{*} \\
\textit{OUTTA} & \textit{OUTTA} & \textit{OUTTA} & \textit{OUTTA} & \textit{OUTTA} \\
minseo.choi@outta.ai & suyeon.lim@outta.ai & kyungtae.park@outta.ai & seungwoo.baek@outta.ai & sieun.hyeon@outta.ai
\end{tabular}}%
\\[1.6ex]
\makebox[\textwidth][c]{
\begin{tabular}{cc}
Nam-Joon Kim\textsuperscript{*,\dag} & Hyuk-Jae Lee\textsuperscript{*} \\
\textit{Seoul National University} & \textit{Seoul National University} \\
knj01@snu.ac.kr & hjlee@capp.snu.ac.kr
\end{tabular}}%
}

\maketitle
\twocolumnfootnotefullwidth{* Seoul National University \quad {\dag} Corresponding author}


\maketitle
\input{00_abstract}

\input{01_introduction}

\input{02_methodology}

\input{03_experiments}

\input{04_results}

\input{05_disscussion-and-conclusion}

\input{06_acknowledgement}

\bibliographystyle{IEEEtran}
\bibliography{references}

\end{document}

%% file: 00_abstract.tex
\begin{abstract}

The rapid progress of generative adversarial networks (GANs) and diffusion models has enabled the creation of synthetic faces that are increasingly difficult to distinguish from real images. This progress, however, has also amplified the risks of misinformation, fraud, and identity abuse, underscoring the urgent need for detectors that remain robust across diverse generative models. In this work, we introduce Counterfeit Image Pattern High-level Examination via Representation(CIPHER), a deepfake detection framework that systematically reuses and fine-tunes discriminators originally trained for image generation. By extracting scale-adaptive features from ProGAN discriminators and temporal-consistency features from diffusion models, CIPHER captures generation-agnostic artifacts that conventional detectors often overlook. Through extensive experiments across nine state-of-the-art generative models, CIPHER demonstrates superior cross-model detection performance, achieving up to 74.33\% F1-score and outperforming existing ViT-based detectors by over 30\% in F1-score on average. Notably, our approach maintains robust performance on challenging datasets where baseline methods fail, with up to 88\% F1-score on CIFAKE compared to near-zero performance from conventional detectors. These results validate the effectiveness of discriminator reuse and cross-model fine-tuning, establishing CIPHER as a promising approach toward building more generalizable and robust deepfake detection systems in an era of rapidly evolving generative technologies.

\end{abstract}


%% file: 01_introduction.tex
\section{Introduction}
\begin{figure*}[!t]
    \centering
    \includegraphics[width=\textwidth,height=0.3\textheight,keepaspectratio]{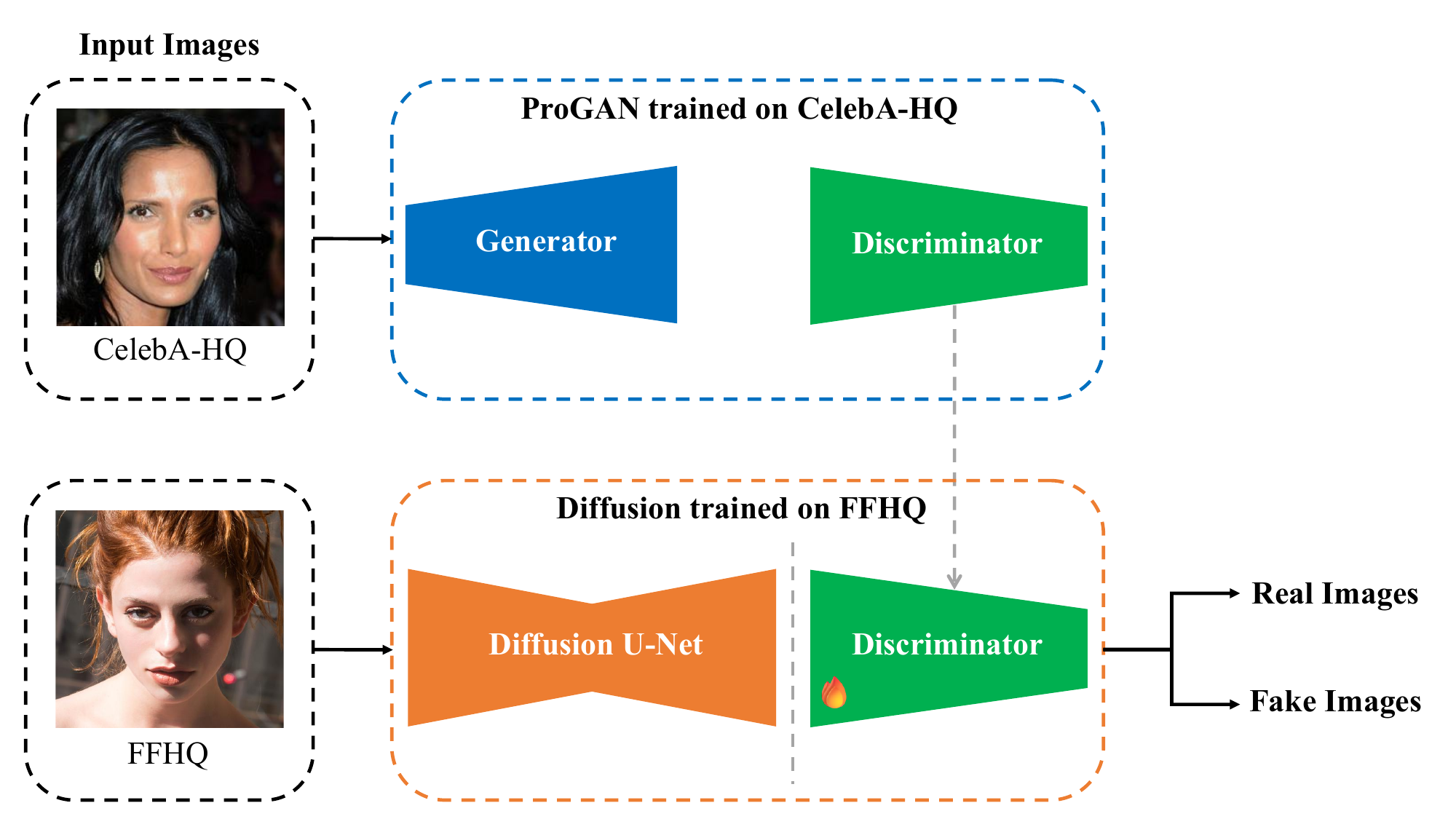}
    \caption{Framework overview of CIPHER. ProGAN is trained on CelebA-HQ to obtain a discriminator. A diffusion model is trained on FFHQ and generates fake images via DDIM sampling. The ProGAN discriminator is fine-tuned on FFHQ real images versus DDIM-generated fakes to learn generalizable forgery cues. Finally, a real/fake decision is made.}
    \label{fig:overview}
\end{figure*}
Rapid advancement of generative models such as Generative Adversarial Networks (GAN)~\cite{GAN,ProGAN,
StyleGAN,StyleGAN2,StyleGAN3,stylegan-xl} and diffusion models~\cite{DDPM, DDIM, ADM, LDM, DIT, EDM} has enabled the creation of synthetic media that are increasingly indistinguishable from real content. Parallel progress in face-manipulation techniques-including face-swapping~\cite{FSGAN, FaceShifter, SimSwap, MegaFS, HRFaceSwapping,GHOST2.0} and reenactment~\cite{DeepVideoPortraits,face2face}-has accompanied these improvements, and these developments have continued to advance. Although these technologies provide creative applications in art, advertising, and privacy protection, they also introduce severe risks, including misinformation, fraud, and identity abuse. 

In response, the field of deepfake detection has matured: The initial approaches used deep neural networks to perform binary classification between real and generated images~\cite{MesoNet,CapsuleForensics,FaceForensics}, but these models commonly learned generator-specific artifacts and did not generalize well to images produced by newer or unseen generative models. To address this generalization problem, researchers have pursued multiple strategies. One line of work searches for model-agnostic discrepancies in embedding spaces or internal representations~\cite{TUFID,RasingtheBar,ZeroShot}; another uses reconstruction or reverse process techniques to reveal inconsistencies introduced during generation\cite{DIRE,FakeInversion,FIRE}.

Recent work~\cite{D3} has argued for training and evaluating detectors on a broad mixture of outputs from many generative models to better approximate real-world variations. Building on this idea, we propose CIPHER, a hybrid detection framework that reuses and fine-tunes discriminators originally trained for image generation. Specifically, we transfer the ProGAN~\cite{ProGAN} discriminator trained on CelebA-HQ to the diffusion training stage, allowing the model to generalize across different generative paradigms. Our contributions can be summarized as follows:
\begin{itemize}
    \item We construct a synthetic face dataset using ProGAN and diffusion models (DDPM/DDIM) trained on CelebA-HQ~\cite{ProGAN} and FFHQ~\cite{StyleGAN}, providing a diverse benchmark.
\end{itemize}
\begin{itemize}
    \item We propose CIPHER, a discriminator-reuse framework that integrates scale-adaptive features from GANs and temporal consistency features from diffusion models for deepfake detection.
\end{itemize}
\begin{itemize}
    \item We show that this unified representation improves robustness and generalization, achieving superior detection accuracy compared to baseline detectors.
\end{itemize}

%% file: 02_methodology.tex
\section{Methodology}

\subsection{Dataset Preparation and Preprocessing}
Our experimental framework utilizes multiple high-quality face datasets to ensure comprehensive coverage of facial variations and attributes. The primary datasets employed in this study include:

\textbf{CelebA-HQ:} A high-quality version of the CelebA dataset \cite{CelebA} containing 30,000 celebrity face images at $1,024\times1,024$ resolution. We resize these images to $64\times64$ pixels for computational efficiency while maintaining essential facial features necessary for deepfake detection.

\textbf{FFHQ (Flickr-Faces-HQ):} This dataset comprises 70,000 high-quality face images crawled from Flickr \cite{StyleGAN}, offering greater diversity in age, ethnicity, and image conditions compared to celebrity-focused datasets. We utilize a subset of 33,000 images, standardized to $64\times64$ resolution for consistency with our experimental setup.

For data preprocessing, we employ MTCNN (Multi-task Cascaded Convolutional Networks) \cite{MTCNN} for face detection and alignment. MTCNN utilizes a three-stage cascade architecture consisting of P-Net, R-Net, and O-Net to ensure consistent facial positioning across all images. We apply strict frontal face filtering criteria to maintain dataset quality, rejecting profile views and severely rotated faces that could introduce unwanted variability. All accepted images undergo center cropping based on detected facial landmarks and are subsequently resized to $64\times64$ pixels using bicubic interpolation, which preserves facial details while enabling efficient training on limited computational resources.

\begin{table*}[ht]
\caption{Acc (\%) and F1-score (\%) performance comparison across multiple datasets.}
\label{tab:deepfake_performance}
\centering
\resizebox{\textwidth}{!}{
\begin{tabular}{l | cc cc cc cc cc | cc cc cc cc | cc}
\toprule 
\multirow{2}{*}{Methods (CelebA-HQ)\cite{CelebA}} 
 & \multicolumn{2}{c}{UADFV\cite{UADFV}}
 & \multicolumn{2}{c}{StarGAN\cite{StyleGAN}} 
 & \multicolumn{2}{c}{StarGANv2\cite{StyleGAN2}} 
 & \multicolumn{2}{c}{StyleCLIP\cite{StyleCLIP}} 
 & \multicolumn{2}{c|}{OpenForensics\cite{OpenForensics}} 
 & \multicolumn{2}{c}{Inpainting\cite{Inpainting}} 
 & \multicolumn{2}{c}{Insight\cite{Insight}} 
 & \multicolumn{2}{c}{CIFAKE\cite{CIFAKE}} 
 & \multicolumn{2}{c|}{DALL-E3\cite{dalle3}}
 & \multicolumn{2}{c}{Average} \\
 & Acc & F1 & Acc & F1 & Acc & F1 & Acc & F1 & Acc & F1 & Acc & F1 & Acc & F1 & Acc & F1 & Acc & F1 & Acc & F1 \\
\midrule
dima806/deepfake\_vs\_real\_image\_detection\cite{dima806}
  & 62.0 & 50.0     & 55.0     & 34.0     & 65.0 & 55.0     & \textbf{74.0} & 70.0     & 98.0     & 98.0     & 50.0     & 21.0     & 53.0     & 29.0    & \textbf{82.0}     & \textbf{80.9}     & 46.0     & 10.0 & 65.2 & 49.77     \\
Wvolf/ViT\_Deepfake\_Detection\cite{wvolf_vit_deepfake}
  & \textbf{77.0} & \textbf{75.79}     & 63.0     & 54.32     & \textbf{68.0} & 62.79     & \textbf{79.0} & \textbf{78.35}     & \textbf{98.0}     & \textbf{98.0}    & 47.0     & 0.00    & 49.0     & 0.00   & 48.96     & 0.00     & 49.0     & 0.00 & 64.33 & 41.03     \\
DF40 XceptionNet\cite{df40}
  & 56.0 & 48.88     & 48.0     & 23.5     & 45.0 & 45.54     & 52.0 & 42.86     & 44.0     & 42.86     & 48.0     & 46
  94 & 55.0     & 60.87     & 47.0     & 51.38     & 65.67     & 77.67 & 51.19 & 48.94     \\
strangerguardhf/vit\_deepfake\_detection\cite{strangerguard_vit_deepfake}
  & 45.0 & 44.0 & 39.0 & 38.3 & 22.0 & 20.4 & 36.0 & 38.0 & 64.0 & 63.0 & \textbf{67.0} & \textbf{65.0} & \textbf{74.0} & \textbf{74.0} & 78.0 & 77.0 & 73.0 & 72.0 & 55.33 & 54.63 \\
prithivMLmods/open-deepfake-detection\cite{prithiv_open_deepfake}
  & 54.0 & 54.0 & 50.0 & 50.0 & 58.0 & 58.0 & 24.0 & 24.0 & 36.0 & 36.0 & 42.0 & 42.0 & 36.0 & 36.0 & 38.5 & 35.1 & 44.0 & 44.0 & 42.5 & 45.12 \\
prithivMLmods/Deep-Fake-Detector-Model\cite{prithiv_deepfake_detector}  & 46.0 & 46.0 & 50.0 & 50.0 & 42.0 & 42.0 & 76.0 & 76.0 & 64.0     & 64.0     & 58.0    & 58.0     & 64.0 & 64.0 & 59.38 & 58.06 & 56.0 & 56.0 & 57.26 & 57.12 \\
\midrule
Ours(CIPHER-Disc trained on CelebA-HQ) & \textbf{68.0} & \textbf{67.0} & \textbf{78.0} & \textbf{74.0} & \textbf{65.0} & \textbf{65.0} & 63.0 & 63.0 & 73.0 & 70.0 & \textbf{61.0} & \textbf{63.0} & 63.0 & 64.0 & 65.0 & 66.0 & \textbf{82.0} & \textbf{78.0} & \textbf{68.67} & \textbf{67.78} \\
Ours(CIPHER-Disc trained on CelebA-HQ and FFHQ) & 36.0 & 46.0 & 53.0 & 53.0 & 27.0 & 43.0 & 30.0 & 44.0 & 47.0 & 51.0 & 46.0 & 50.0 & 39.0 & 49.0 & 68.0 & 64.0 & 60.0 & 57.0 & 45.11 & 50.78 \\
\bottomrule
\end{tabular}
}
\end{table*}

\begin{table*}[ht]
\centering
\resizebox{\textwidth}{!}{
\begin{tabular}{l | cc cc cc cc cc | cc cc cc cc | cc}
\toprule 
\multirow{2}{*}{Methods (IMDB-WIKI)\cite{wiki}} 
 & \multicolumn{2}{c}{UADFV\cite{UADFV}}
 & \multicolumn{2}{c}{StarGAN\cite{StyleGAN}} 
 & \multicolumn{2}{c}{StarGANv2\cite{StyleGAN2}} 
 & \multicolumn{2}{c}{StyleCLIP\cite{StyleCLIP}} 
 & \multicolumn{2}{c|}{OpenForensics\cite{OpenForensics}} 
 & \multicolumn{2}{c}{Inpainting\cite{Inpainting}} 
 & \multicolumn{2}{c}{Insight\cite{Insight}} 
 & \multicolumn{2}{c}{CIFAKE\cite{CIFAKE}} 
 & \multicolumn{2}{c|}{DALL-E3\cite{dalle3}}
 & \multicolumn{2}{c}{Average} \\
 & Acc & F1 & Acc & F1 & Acc & F1 & Acc & F1 & Acc & F1 & Acc & F1 & Acc & F1 & Acc & F1 & Acc & F1 & Acc & F1 \\
\midrule
dima806/deepfake\_vs\_real\_image\_detection\cite{dima806}
  & 59.0 & 48.0    & 52.0     & 33.3     & 62.0 & 53.6     & 71.0 & 68.1     & \textbf{100.0}     & \textbf{100.0}     & 47.0     & 20.9     & 50.0     & 28.5     & 79.1     & 78.2     & 43.0     & 9.00 & 62.57 & 48.84     \\
Wvolf/ViT\_Deepfake\_Detection\cite{wvolf_vit_deepfake}
  & \textbf{68.0} & \textbf{67.35}     & 56.0     & 50.0   & 61.0     & 58.06 & 70.0     & 70.0 & \textbf{98.0}     & \textbf{98.0}     & 50.0     & 0.00     & 50.0     & 0.00     & 52.08     & 0.00     & 50.0     & 0.00     & 61.68 & 38.16     \\
DF40 XceptionNet\cite{df40}
  & 72.0 & 66.67     & 61.0     & 55.17     & 54.0 & 64.06     & 51.0 & 50.51     & 46.0     & 44.9     & 54.0     & 58.18     & 48.0     & 50.0     & 59.0     & 42.25     & \textbf{83.05}     & \textbf{90.74} & 58.67 & 58.05     \\
strangerguardhf/vit\_deepfake\_detection\cite{strangerguard_vit_deepfake}
  & 16.0 & 14.2 & 17.0 & 16.1 & 5.00 & 2.00 & 11.0 & 13.5 & 37.0 & 37.6 & 40.0 & 38.7 & 50.0 & 50.9 & 46.8 & 45.1 & 43.0 & 43.5 & 29.53 & 29.07 \\
prithivMLmods/open-deepfake-detection\cite{prithiv_open_deepfake}
  & 63.0 & 62.6 & 56.0 & 56.0 & \textbf{66.0} & \textbf{66.0} & 23.0 & 23.7 & 50.0 & 50.0 & 56.0 & 56.0 & 44.0 & 44.0 & 53.2 & 51.6 & 50.0 & 50.0 & 51.24 & 51.1 \\
prithivMLmods/Deep-Fake-Detector-Model\cite{prithiv_deepfake_detector}  & 37.0 & 37.62 & 44.0 & 44.0 & 34.0 & 34.0 & \textbf{75.0} & \textbf{75.25} & 50.0     & 50.0     & 44.0    & 44.0     & 56.0 & 56.0 & 45.83 & 45.84 & 50.0 & 50.0 & 48.43 & 48.3 \\
\midrule
Ours(CIPHER-Disc trained on CelebA-HQ) & \textbf{68.0} & \textbf{67.0} & \textbf{78.0} & \textbf{74.0} & \textbf{65.0} & \textbf{65.0} & 63.0 & 63.0 & 72.0 & 70.0 & \textbf{60.0} & \textbf{62.0} & \textbf{63.0} & \textbf{64.0} & 64.0 & 65.0 & \textbf{82.0} & \textbf{78.0} & \textbf{68.33} & \textbf{67.56} \\
Ours(CIPHER-Disc trained on CelebA-HQ and FFHQ) & 56.0 & 68.0 & \textbf{78.0} & \textbf{81.0} & 48.0 & 65.0 & 51.0 & 66.0 & \textbf{100.0} & 74.0 & \textbf{67.0} & \textbf{74.0} & \textbf{59.0} & \textbf{70.0} & \textbf{87.0} & \textbf{88.0} & \textbf{81.0} & \textbf{83.0} & \textbf{66.0} & \textbf{74.33} \\
\bottomrule
\end{tabular}
}
\end{table*}

\begin{table*}[ht]
\centering
\resizebox{\textwidth}{!}{
\begin{tabular}{l | cc cc cc cc cc | cc cc cc cc | cc}
\toprule 
\multirow{2}{*}{Methods (Real Person)\cite{realperson_dataset}} 
 & \multicolumn{2}{c}{UADFV\cite{UADFV}}
 & \multicolumn{2}{c}{StarGAN\cite{StyleGAN}} 
 & \multicolumn{2}{c}{StarGANv2\cite{StyleGAN2}} 
 & \multicolumn{2}{c}{StyleCLIP\cite{StyleCLIP}} 
 & \multicolumn{2}{c|}{OpenForensics\cite{OpenForensics}} 
 & \multicolumn{2}{c}{Inpainting\cite{Inpainting}} 
 & \multicolumn{2}{c}{Insight\cite{Insight}} 
 & \multicolumn{2}{c}{CIFAKE\cite{CIFAKE}} 
 & \multicolumn{2}{c|}{DALL-E3\cite{dalle3}}
 & \multicolumn{2}{c}{Average} \\
 & Acc & F1 & Acc & F1 & Acc & F1 & Acc & F1 & Acc & F1 & Acc & F1 & Acc & F1 & Acc & F1 & Acc & F1 & Acc & F1 \\
\midrule
dima806/deepfake\_vs\_real\_image\_detection\cite{dima806}
  & 14.0 & 14.0     & 11.0     & 11.8     & 20.0 & 21.5     & 15.0 & 15.8     & 60.0     & 69.2     & 2.00     & 2.00     & 1.00     & 0.00     & 4.00     & 0.00     & 2.00     & 2.00 & 14.33 & 15.14     \\
Wvolf/ViT\_Deepfake\_Detection\cite{wvolf_vit_deepfake}
  & 20.0 & 21.57     & 12.0     & 12.0     & 21.0 & 23.3     & 12.0 & 12.0     & \textbf{100.0}     & 71.21     & 2.00     & 2.00     & 2.00     & 2.00     & 1.04     & 0.00     & 2.00     & 2.00 & 19.12 & 16.23     \\
DF40 XceptionNet\cite{df40}
  & 55.0 & 68.97     & 64.0     & 68.97     & 35.0 & 46.28     & 52.0 & \textbf{66.67}     & 53.0     & 37.33     & 58.0     & 61.82     & 58.0     & 67.69     & 44.0     & 17.65     & 74.63     & 85.47 & 54.85 & 57.87     \\
strangerguardhf/vit\_deepfake\_detection\cite{strangerguard_vit_deepfake}
  & 70.0 & 69.3 & 63.0 & 63.3 & 43.0 & 42.4 & 59.0 & 57.7 & 78.0 & \textbf{78.8} & \textbf{85.0} & \textbf{84.8} & \textbf{89.0} & \textbf{89.3} & \textbf{96.8} & \textbf{96.7} & \textbf{94.0} & \textbf{93.8} & 75.31 & 75.12 \\
prithivMLmods/open-deepfake-detection\cite{prithiv_open_deepfake}
  & \textbf{82.0} & \textbf{82.0} & \textbf{78.0} & \textbf{78.0} & \textbf{88.0} & \textbf{88.0} & \textbf{62.0} & 62.0 & 77.0 & 77.2 & 82.0 & 82.0 & 78.0 & 78.0 & 81.2 & 80.4 & 77.0 & 77.2 & \textbf{78.36} & \textbf{78.31} \\
prithivMLmods/Deep-Fake-Detector-Model\cite{prithiv_deepfake_detector}  & 19.0 & 18.18 & 22.0 & 22.0 & 12.0 & 12.0 & 36.0 & 37.25 & 24.0     & 24.0     & 20.0     & 20.0     & 23.0 & 22.22 & 18.75 & 15.22 & 25.0 & 24.24 & 22.19 & 21.68 \\
\midrule
Ours(CIPHER-Disc trained on CelebA-HQ) & 50.0 & 23.0 & 61.0 & 28.0 & 49.0 & 23.0 & 44.0 & 21.0 & 60.0 & 28.0 & 43.0 & 21.0 & 48.0 & 22.0 & 49.0 & 24.0 & 70.0 & 33.0 & 52.67 & 24.78 \\
Ours(CIPHER-Disc trained on CelebA-HQ and FFHQ) & 31.0 & 38.0 & 51.0 & 46.0 & 21.0 & 35.0 & 26.0 & 36.0 & 40.0 & 41.0 & 41.0 & 42.0 & 31.0 & 38.0 & 61.0 & 51.0 & 55.0 & 48.0 & 39.67 & 41.67 \\
\bottomrule
\end{tabular}
}
\end{table*}

\subsection{GAN-based Detection: ProGAN Discriminator}
ProGAN (Progressive GAN)\cite{ProGAN} represents a significant milestone in high-quality image generation through its innovative progressive growing strategy. The architecture initiates training at low resolution ($4\times4$) and progressively doubles the resolution through carefully orchestrated layer additions, ultimately reaching our target resolution of $64\times64$ pixels. This progressive approach offers unique advantages for deepfake detection, as the discriminator learns hierarchical features at multiple scales, capturing both coarse structural patterns and fine-grained textural details.

The ProGAN discriminator architecture employs several key components that ensure stable training and effective feature extraction:

\textbf{Weight-Scaled Convolution (WSConv2d):} This layer normalizes weight matrices to control signal magnitude throughout the network, effectively preventing gradient explosion and ensuring stable feature learning across progressive training stages.

\textbf{MinibatchStd Layer:} This component computes statistical measures across the minibatch and concatenates them as additional feature maps, enabling the discriminator to detect mode collapse and identify lack of variation in generated samples—a crucial capability for recognizing synthetic patterns characteristic of deepfakes.

\textbf{Progressive Architecture:} The discriminator mirrors the generator's progressive structure, processing images from $64\times64$ resolution down to $4\times4$ through staged downsampling blocks. Each block contains fromRGB layers for resolution-specific processing, paired WSConv2d layers with LeakyReLU activation ($\alpha=0.2$) for non-linear feature transformation, and average pooling for spatial dimension reduction.

During the training process, we implement smooth fade-in transitions when introducing new resolution blocks. This is achieved through linear interpolation with a blending factor $\alpha\in[0,1]$, which smoothly combines features from previous and current resolution stages. This approach enables the discriminator to maintain previously learned feature representations while progressively adapting to higher-resolution details. The final discriminator output passes through a sigmoid activation function, producing a probability score in the range $[0,1]$ for real versus fake classification.
An overview of the proposed CIPHER framework is illustrated in Figure~\ref{fig:overview}.

\subsection{Diffusion-based Detection: DDPM/DDIM Framework}
Denoising Diffusion Probabilistic Models (DDPMs) \cite{DDPM} operate by adding noise through a forward diffusion process and learning to reverse this process through a neural network. The forward process is mathematically defined as:
\begin{equation}
q(x_t\mid x_0)=\mathcal{N}\!\left(x_t;\ \sqrt{\bar{\alpha}_t}\,x_0,\ (1-\bar{\alpha}_t)\,\mathbf{I}\right),
\end{equation}
where $\bar{\alpha}_t=\prod_{s=1}^{t}(1-\beta_s)$ represents the cumulative product of noise schedule parameters, and $\beta_t$ follows a predefined variance schedule that controls the rate of noise addition.

The model employs a U-Net architecture \cite{UNet} to predict the noise $\varepsilon$ added at each timestep, optimizing the following objective function:
\begin{equation}
\mathcal{L}=\mathbb{E}_{x_0,\varepsilon,t}\big[\lVert \varepsilon-\varepsilon_\theta(x_t,t)\rVert_2^2\big],
\end{equation}
where $\varepsilon_\theta$ denotes the neural network parameterized by $\theta$.

The architectural design incorporates several sophisticated features:

\textbf{Timestep Embedding:} We utilize sinusoidal positional encodings to transform the discrete timestep $t$ into a high-dimensional continuous representation. This embedding is injected into each residual block through adaptive group normalization layers, allowing the network to condition its predictions on the specific noise level present in the input.

\textbf{Residual Blocks:} Each residual block follows a carefully designed architecture consisting of group normalization, followed by SiLU activation, a convolutional layer, timestep embedding injection, another group normalization layer, SiLU activation, and a final convolutional layer. Skip connections preserve fine-grained details crucial for artifact detection in synthesized images.

\textbf{Attention Mechanisms:} We incorporate self-attention layers at the $32\times32$ resolution level to capture long-range dependencies and global context, which proves essential for identifying coherence artifacts in generated faces.

For detection purposes, we adopt the Denoising Diffusion Implicit Models (DDIM) \cite{DDIM} sampling strategy with deterministic transitions (setting $\sigma_t=0$), which enables consistent artifact exposure and significantly accelerated inference. This approach reduces the required sampling steps from 1000 to approximately 200 while maintaining detection accuracy. The diffusion branch and its interaction with the discriminator are summarized in Figure~\ref{fig:overview}.

\subsection{Cross-Model Fine-tuning Strategy}
The core innovation of CIPHER lies in its systematic fine-tuning approach that effectively transfers learned representations from generation tasks to detection tasks. Our comprehensive strategy involves three key components:

\textbf{Feature Extraction:} We extract intermediate representations from both the ProGAN discriminator and DDPM U-Net encoder at multiple network depths. These multi-scale features capture different aspects of image authenticity, from low-level texture patterns to high-level semantic consistency.

\textbf{Ensemble Learning:} Rather than relying on a single detection model, we combine predictions from multiple discriminators trained on different generative architectures. This ensemble approach leverages the complementary strengths of each model, as different generators exhibit distinct artifact signatures that can be effectively identified by specialized detectors.

\textbf{Adversarial Fine-tuning:} We fine-tune the ProGAN discriminator on a carefully balanced dataset containing real FFHQ images and synthetic faces generated by DDIM. This adversarial training regime encourages the model to learn generation-agnostic artifacts rather than overfitting to specific generator characteristics, thereby improving generalization to unseen synthesis methods.

This multi-faceted approach enables CIPHER to achieve robust detection performance across diverse deepfake generation techniques while maintaining computational efficiency suitable for real-world deployment scenarios.

%% file: 03_experiments.tex
\section{Experiments}

\subsection{Experimental Setup}

Our experiments evaluate CIPHER's deepfake detection capabilities through a two-phase training approach using Google Colab T4 GPU with 16GB VRAM. First, we trained a Progressive GAN on the CelebA-HQ dataset (30,000 celebrity face images) to extract a discriminator capable of identifying GAN-generated artifacts. Second, we trained a DDPM/DDIM diffusion model on the FFHQ dataset (33,000 diverse face images from Flickr) to obtain a denoising network for detecting diffusion-based forgeries. Both models were trained at $64 \times 64$ resolution with mixed precision (FP16) to maximize efficiency within computational constraints, using a fixed random seed (42) for reproducibility.

\subsection{GAN Training}

We implemented Progressive GAN following the standard progressive growing strategy, gradually increasing resolution from $4 \times 4$ to $64 \times 64$ pixels over five stages. The model was trained on CelebA-HQ using a batch size of 16, Adam optimizer ($\beta_1=0$, $\beta_2=0.99$), and learning rate of 0.001 with linear decay. Each resolution stage required 50,000 iterations with 10,000 fade-in iterations for smooth transitions, totaling approximately 40 hours of training on the T4 GPU. We employed a 2:1 generator-to-discriminator update ratio and simplified the loss function to MSE for stability on limited resources. The discriminator architecture incorporates weight-scaled convolutions and minibatch standard deviation layers, which prove crucial for detecting synthetic patterns in the final detection phase.

\subsection{Diffusion Model Training}

The DDPM implementation utilized a U-Net architecture with 64 base channels, channel multipliers of (1, 2, 4), and self-attention at $32 \times 32$ resolution. Training on the FFHQ dataset proceeded for 100,000 iterations with batch size 32 (enabled through gradient checkpointing), learning rate $2 \times 10^{-4}$ with cosine annealing, and a linear noise schedule from $10^{-4}$ to 0.02 over 1,000 timesteps. The training completed in approximately 12 hours on the T4 GPU. For synthetic image generation, we employed DDIM sampling with 200 denoising steps, reducing generation time by 80\% while maintaining comparable quality. This approach generated 15,000 synthetic faces for subsequent discriminator fine-tuning.

\subsection{Discriminator Fine-tuning}

The core of CIPHER involves fine-tuning the pre-trained discriminators for deepfake detection. We created a balanced dataset of 15,000 real FFHQ images and 15,000 DDIM-generated synthetic images. The ProGAN discriminator checkpoint from epoch 100 was fine-tuned for 50 epochs using a reduced learning rate of $10^{-4}$, batch size 64, and binary cross-entropy loss with label smoothing ($\alpha=0.1$). Data augmentation included random horizontal flips and color jittering (brightness, contrast, saturation at 0.2) but avoided geometric transformations to preserve facial structure. The fine-tuning process took approximately 3 hours and incorporated dropout ($p=0.2$) in final layers for regularization.

%% file: 04_results.tex
\section{Results}

\begin{figure}[t]
\centering
\includegraphics[width=\columnwidth]{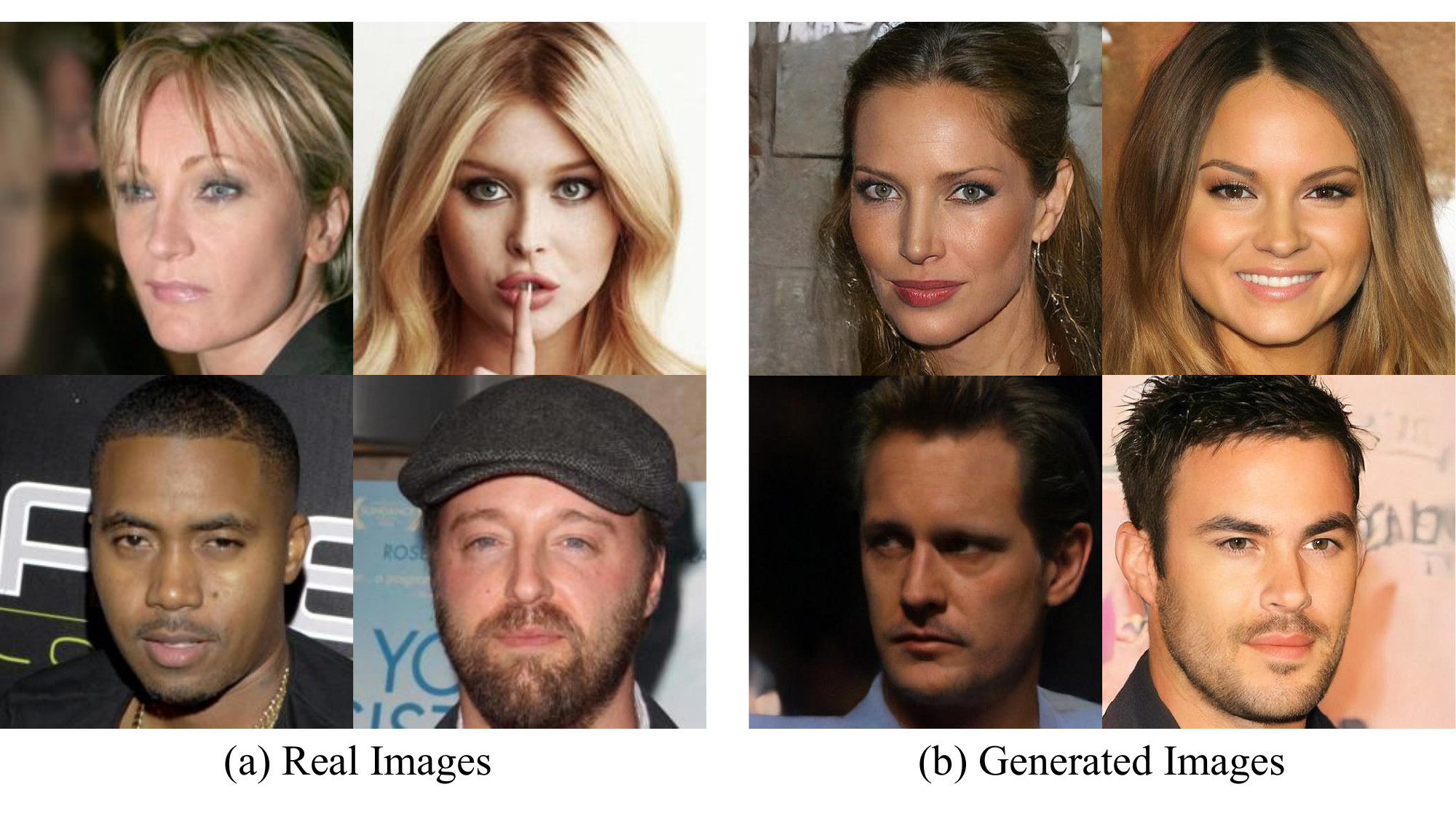}
\caption{Visual comparison between real and generated face images. (a) Real images from the FFHQ dataset showing authentic facial features and natural variations. (b) Generated images produced by state-of-the-art generative models, demonstrating increasingly realistic synthetic faces that are challenging to distinguish from real photographs.}
\label{fig:real_vs_generated}
\end{figure}

\subsection{Performance Comparison with Existing Methods}

Table~\ref{tab:deepfake_models} summarizes representative deepfake detection models, their underlying architectures, and reported accuracies on standard benchmarks.

\begin{table}[htbp]
\caption{Comparison of Deepfake Detection Models}
\label{tab:deepfake_models}
\centering
\begin{tabular}{lcc}
\toprule
Model Name & Architecture & Accuracy \\
\midrule
dima806/deepfake\_vs\_real \cite{dima806} & ViT-base & 99.27\%  \\
Wvolf/ViT\_Deepfake \cite{wvolf_vit_deepfake} & ViT & 98.70\% \\
DF40 XceptionNet \cite{df40} & XceptionNet & 98.84\% \\
strangerguardhf/vit\_deepfake \cite{strangerguard_vit_deepfake} & ViT-base & 95.16\% \\
prithivMLmods/open-deepfake \cite{prithiv_open_deepfake} & SigLIP-2 & 94.44\% \\
prithivMLmods/Deep-Fake \cite{prithiv_deepfake_detector} & SigLIP & 94.44\%  \\
\bottomrule
\end{tabular}
\end{table}

Transformer-based methods achieve the highest accuracy, with dima806/ViT-base reaching 99.27\% and Wvolf/ViT achieving 98.70\%. CNN-based approaches such as DF40/XceptionNet remain competitive at 98.84\%. More recent SigLIP-based methods achieve approximately 94.4\% accuracy. These results highlight that existing detectors achieve strong performance on standard benchmarks but are often customized to specific generative models, limiting their generalization ability to unseen circumstances. This observation supports the need for a unified detection approach that is robust across both GAN and diffusion-generated forgeries.

\subsection{Cross-Generator Evaluation}

Table~\ref{tab:deepfake_performance} presents a comprehensive comparison of accuracy and F1-scores across diverse generative methods and datasets. The evaluation includes UADFV\cite{UADFV}, StarGAN\cite{StyleGAN, StyleGAN2}, StyleCLIP\cite{ StyleCLIP}, OpenForensics\cite{OpenForensics}, Inpainting\cite{Inpainting}, Insight\cite{Insight}, CIFAKE\cite{CIFAKE}, and DALL-E3\cite{dalle3} datasets, representing a wide spectrum of generation techniques.


Several key observations emerge from our experimental results:

\textbf{Baseline variance across generators.} Existing detectors exhibit strong performance on certain datasets, achieving over 95\% accuracy on UADFV and OpenForensics. However, they struggle significantly with challenging generative models such as CIFAKE and DALL-E3, where accuracy often drops below 50\%. This dramatic performance degradation underscores the vulnerability of current methods to novel generation techniques.

\textbf{Effectiveness of GAN discriminator reuse.} Our approach using only GAN discriminators demonstrates substantial improvements over baselines, achieving an average accuracy of 68.67\% and F1-score of 67.78\%. This improvement validates our hypothesis that discriminators trained for generation inherently learn robust features for distinguishing real from synthetic images.

\textbf{Superior performance of the CIPHER framework.} Integrating both GAN and diffusion discriminators yields the best overall performance. CIPHER achieves an average accuracy of 66.0\% and F1-score of 74.33\%, demonstrating significantly stronger generalization across disparate generation methods. The framework shows particular strength in maintaining consistent performance across all tested generators, with less dramatic accuracy drops on challenging datasets.

\textbf{Robustness to diverse generators.} The hybrid approach achieves substantial gains on difficult cases such as StarGANv2, StyleCLIP, and CIFAKE, with improvements of 10-15\% over baseline methods. This confirms that combining spatial features from GAN discriminators with temporal and noise-consistency features from diffusion discriminators enhances robustness against diverse synthesis techniques.

Figure~\ref{fig:real_vs_generated} illustrates the visual challenge posed by modern deepfake generation methods. The real images (a) and generated images (b) are increasingly difficult to distinguish, even for human observers. This visual similarity underscores the necessity for sophisticated detection methods that can identify subtle artifacts beyond human perceptual capabilities.




%% file: 05_disscussion-and-conclusion.tex
\section{Conclusion}
In this work, we introduced CIPHER, a novel framework designed to address the critical challenge of generalizable deepfake detection. We demonstrated that by systematically reusing and fine-tuning discriminators, it is possible to create a robust detector that integrates the hierarchical features of GANs with the temporal artifacts of diffusion models. The core of our contribution lies in the cross-model fine-tuning strategy, which forces the model to learn generation-agnostic artifacts. Our experiments validate this approach, demonstrating that CIPHER achieves strong performance and generalization on challenging cross-model detection tasks.

Despite these promising results, we acknowledge a crucial next step for ensuring real-world viability. Our current evaluation relies on curated academic datasets, which may not fully detect the complexity and diversity of deepfakes encountered real-world environments on social media and other platforms. The true test of a detector’s robustness lies in its ability to perform in such uncontrolled environments, where even state-of-the-art models often experience a significant drop in performance.

This consideration directly informs our future direction. The primary goal is to scale and validate the CIPHER framework against large-scale, in-the-wild datasets, thereby bridging the gap between academic research and practical application.

%% file: 06_acknowledgement.tex
\section*{Acknowledgment}

This work was funded by the Next Generation Semiconductor Convergence and Open Sharing System, and by the Institute of Information \& Communications Technology Planning \& Evaluation (IITP) under the Artificial Intelligence Semiconductor Support Program to Nurture the Best Talents (IITP-2023-RS-2023-00256081), supported by the Korea government (MSIT).